\documentclass[10pt,onecolumn,letterpaper]{article}

\usepackage{wacv}
\usepackage{times}
\usepackage{epsfig}
\usepackage{graphicx}
\usepackage{amsmath}
\usepackage{amssymb}
\wacvfinalcopy

\ifwacvfinal\fi
\begin{document}

\title{CT-SRCNN: Cascade Trained and Trimmed Deep Convolutional Neural Networks for Image Super Resolution}

\author{Haoyu Ren \hspace{2cm} Mostafa El-Khamy \hspace{2cm} Jungwon Lee \\
SOC R\&D, Samsung Semiconductor Inc. \\
San Diego, California, USA \\
{\tt\small \{haoyu.ren, mostafa.e, jungwon2.lee\}@samsung.com}
}

\maketitle
\ifwacvfinal\thispagestyle{empty}\fi

\begin{abstract}
We propose methodologies to train highly accurate and efficient deep convolutional neural networks (CNNs) for image super resolution (SR). A cascade training  approach to deep learning is proposed to improve the accuracy of the neural networks while gradually increasing the number of network layers. Next, we 
explore how to improve the SR efficiency by making the network slimmer. Two methodologies, the one-shot trimming and the cascade trimming, are proposed. With the cascade trimming, the network's size is gradually reduced layer by layer, without significant loss on its discriminative ability. Experiments on benchmark image datasets show that our proposed SR network achieves the state-of-the-art super resolution accuracy, while being more than 4 times faster compared to existing deep super resolution networks. 
\end{abstract}

\section{Introduction}
\noindent{}Super resolution (SR) aims to retrieve a high resolution (HR) image from a given low resolution (LR) image. SR is widely used in many computer vision applications such as surveillance, face and iris recognition, and medical image processing. There have been many SR algorithms proposed recently \cite{nasrollahi2014super, yang2014single}. Some early methods referred to interpolations, statistical image priors \cite{glasner2009super, kim2010single} or contour features \cite{li2001new}. Later, the example-based methods were proposed based on learning a dictionary of patch correspondences, such as neighbor embedding \cite{chang2004super, bevilacqua2012low} and sparse coding \cite{timofte2013anchored, timofte2014a+, yang2010image, timofte2016seven}. Recently, convolutional neural networks (CNNs) have been widely used due to their significant improvement of the accuracy.

The pioneering work SRCNN \cite{dong2016image} introduced CNN as a solution to the SR problem. The accuracy of SRCNN is limited due to the 3-layer structure and its small context receptive field. To improve the accuracy, some researchers propose to use more layers \cite{kim2016accurate}, deep recursive structure \cite{kim2016deeply, tai2017image}, or other network architectures such as ResNet \cite{ledig2016photo, dahl2017pixel}. However, most state of art methods could not be executed in real-time using practical hardware due to their large network size. In addition, with deeper networks, it becomes more difficult to tune the training parameters, such as the weight initialization, the learning rate, and the weight decay rate. As a result, the training can be stuck into a local minimum or does not converge at all. Hence, increasing depth might lead to accuracy saturation \cite{dong2016image} or degradation for image classification \cite{ glasner2009super}.

In this paper, we propose the Cascade Trained Super Resolution Convolutional Neural Network (CT-SRCNN) as a solution to the above problem. Different from existing approaches that train all the layers at once with unsupervised weight initialization, we start by training a small network of 3 layers. When the rate of decrease in the training loss diminishes,  newly initialized layers are gradually inserted into the trained network to make it deeper, and the training is continued. With this cascade training strategy, the convergence is improved, and the accuracy could be consistently increased with more layers. So it is `the deeper the better'. In addition, all the weights of the new layers in the CT-SRCNN are randomly initialized, and the learning rate is fixed. This is a great advantage compared to existing approaches \cite{kim2016accurate, ledig2016photo, dahl2017pixel}, which need much time spent on tuning the parameters. When the CT-SRCNN depth becomes 19 layers, the accuracy is competitive compared to the state-of-the-art image SR networks, while using only $30\%$ of the parameters. The speed is also more than 4 times faster.

Next, we explore how to remove the redundant filters from the SR network to improve the efficiency. We first discuss a one-shot trimming scheme, where all layers are made slimmer by removing filters with less importance. Then a cascade trimming method is proposed, which results in the same slim network but less accuracy loss compared to the one-shot trimming. Similar to cascade training, cascade trimming also consists of several stages. At each trimming stage, some layers are made slimmer by removing some of their filters, while keeping the other layers intact, and is followed by fine-tuning the network. Different layers are trimmed at different stages in a cascade manner. Our trimming strategy is different and much simpler than training a whole newly initialized thin network as in FitNets \cite{Romero2015}. Our cascade trimming methodology is also very different from existing pruning algorithms \cite{Han2016, Choi2017}, where some parameters are set to zero but keeping layer sizes the same. For the same number of non-zero parameters, trimmed networks have lower computational complexity, compressed network size, and are more hardware friendly than pruned networks which tend to have random pruning patterns.  We show that by using cascade trimming together with cascade training we can train an SR network with concurrently high accuracy, good computational efficiency, and significantly smaller size.

As summary, our contributions are two folds:

\noindent{}1. A cascade training strategy is utilized. Cascade training does not need much work on parameter tuning. The resulting CT-SRCNN can achieve the state-of-the-art accuracy at 19 layers, with more than 4 times less complexity compared to existing deep CNN models;

\noindent{}2. A network trimming scheme to reduce the SR network size is proposed. The trade-off between network size and accuracy is explored by the proposed cascade trimming. The redundant filters could be removed with minor loss on the accuracy;

\section{Advances in Super Resolution}
\noindent{}Numerous research addressed the image super resolution problem \cite{nasrollahi2014super, yang2014single}. Some early SR algorithms refer to filtering approaches, such as bilinear, bicubic, and Lanczos filtering. These filtering algorithms may generate smooth outputs without recovering any high-frequency information. To solve this issue, some researchers utilize edge features. Allebach et al. \cite{allebach1996edge} generated a high resolution edge map by first filtering with a rectangular center-on-surround-off filter and then performing piecewise linear interpolation between the zero crossings in the filter output. Li et al. \cite{li2001new}  applied the interpolation constrained  by the geometric duality between the low-resolution covariance and the high-resolution covariance. These approaches are computationally efficient, but the accuracy is limited because they oversimplify the SR problem.

Other approaches assumed a kind of mapping between the LR space and the HR space. Such mapping could be learned from a large number of LR-HR pairs. Early methods refer to Markov network or neighbor search strategies \cite{freeman2002example, chang2004super}. The sparse coding dictionary-based image representation \cite{yang2010image} became popular, where a sparse vector is shared between the LR space and the HR space. Zhang et al. \cite{zhang2012multi} designed a multi-scale dictionary to capture redundancies of similar image patches at different scales to enhance visual effects. Timofte et al. \cite{timofte2013anchored}
anchored the neighborhood embedding of a low resolution patch to the nearest atom in the dictionary and precomputed the corresponding embedding matrix. This approach was further enhanced  by utilizing the training data in the testing procedure \cite{timofte2014a+}. Kim and Kwon \cite{kim2010single} learned a general mapping between the LR and HR spaces from example pairs using kernel ridge regression. Such regression was also solved by random forests \cite{schulter2015fast}. In consideration of the self-similarity, Glasner et al. \cite{glasner2009super} unified the example-based SR and classic multi-image SR by exploiting patch redundancies across scales. Huang et al. \cite{huang2015single} expanded the internal patch search space by allowing geometric variations, explicitly localizing planes in the scene, and using the detected perspective geometry to guide the patch search process. 

Recently, CNNs have been widely adopted for image SR. Dong et al. \cite{dong2016image} trained a 3-layer CNN from the bicubic upsampled LR image to the HR image. Some researchers propose to use more complicated networks. Wang et al. \cite{wang2015deep} integrated a sparse representation prior with a feed-forward network based on the learned iterative shrinkage and thresholding algorithm. Kim et al. \cite{kim2016accurate} increased the number of layers to 20 and used small filters with a high learning rate and an adjustable gradient-clipping. The performance of deeper CNN has been further enhanced by using recursive structures and 
skip connections \cite{kim2016deeply, tai2017image}. Dahl et al. \cite{dahl2017pixel} combined the ResNet with a pixel recursive super resolution architecture, which showed promising results on face and bed SR. Lim et al. \cite{lim2017enhanced} removed unnecessary modules in conventional residual networks, and expanded the model size while stabilizing the training procedure. Others prefer to use perception loss instead of the Mean Square Error (MSE), which is claimed to be closer to natural texture and human vision. Sonderby et al. \cite{sonderby2016amortised} introduced a method for amortized MAP inference, which calculated the MAP estimation directly using CNN. Johnson et al. \cite{johnson2016perceptual} proposed the use of perceptual loss functions for training feed-forward networks for image transformation tasks. Ledig et al. \cite{ledig2016photo} employed a very deep residual network (ResNet), and further presented the Super Resolution Generative Adversarial Network (SRGAN) to obtain HR images with texture similar to natural texture.  These works improve the accuracy of the SR system, at the cost of more layers and parameters and with more difficult hyper-parameter tuning. Other works addressed different strategies for fusion of multiple SR-CNNs to get a fused SR-CNN with superior performance \cite{Ren2017Image}.

Other researchers focused on improving the efficiency by extracting the feature maps in LR space and learning the upscaling filters. Shi et al. \cite{shi2016real} introduced an efficient sub-pixel convolutional layer which learned an array of upscaling filters to upscale the LR feature maps into the HR output. Dong et al. \cite{dong2016accelerating} re-designed the SRCNN by adding smaller filters, deconvolution layers, and feature space shrinkage to accelerate the speed without losing the accuracy. Lai et al. \cite{lai2017deep} proposed the Laplacian pyramid super resolution network (LapSRN) to progressively reconstruct the sub-band residuals of high-resolution images. The deconvolution layers are utilized to upscale the feature maps between different scales. These methods use the upscaling layer instead of initial bicubic upsampling, so the patch size and context receptive field will be small. As a result, the accuracy is relatively lower compared to the methods extracting feature maps from the upsampled LR space.

Most of the above SR networks apply the one-shot training with careful tuning of the network learning hyper-parameters on specific datasets. In contrast, our CT-SRCNN could be trained deeper to achieve a higher accuracy without specific parameter tuning. This is an advantage when generalizing to other datasets. In addition, this is the first paper which discusses network trimming and its application to super resolution. The proposed cascade trimming technique is also novel and is more desirable to existing network pruning strategies. 

\section{CT-SRCNN}
\subsection{Cascade training methodology}
\noindent{}Our SR network takes an interpolated LR image (to the desired size) as input and predicts an HR image as its output. The training and the hyper-parameter tuning become more difficult for deeper networks. A feasible way is to separate the whole training into several cascaded stages, and proceed one by one. This is inspired by the stage-by-stage strategy in conventional machine learning algorithms. For example, in AdaBoost, after each strong classifier is learned, more difficult negative samples are bootstrapped to train the next strong classifier. 

We utilize the cascade strategy in training CNN for super resolution, which is called cascade-trained super resolution convolutional neural network (CT-SRCNN). The first stage of the training starts from 3 layers. When the training error of  the network at current stage stops decreasing significantly, we pause the training and insert new randomly initialized layer(s) into the network as intermediate layers, while inheriting weights of other layers from previous stage. Then the training resumes at next stage. This procedure is recursively applied to make the network deeper and deeper. Since most of the weights are inherited from the previous stage, the convergence will be relatively easy without need for further hyper-parameter tuning.

\begin{figure*}
\centering
\includegraphics[height=3cm]{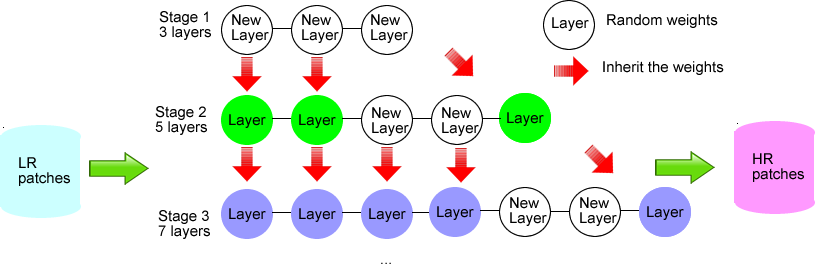}
\caption{Implementation of CT-SRCNN cascade training. In each stage, two new layers with random  weights are inserted.}
\label{fig:example}
\end{figure*}

\subsection{Cascade training implementation}
\noindent{}Let $\mathbf{x}$ denote a LR image and $y$ denote its corresponding HR image. Given a training set $\{(\mathbf{x}_i, y_i), i = 1,\ldots,N\}$ with $N$ samples, our goal is to learn a model $g$ that predicts the  HR output $\hat{y_i} = g(\mathbf{x}_i)$. The training aims to minimize the mean square error (MSE) $\frac{1}{2}\sum_{i=1}^N||y_i-\hat{y}_i||^2$ over the training set.

We start the training from a 3-layer model. The first layer consists of 64 $9\times9$ filters. The second and the third layer consist of 32 $5\times5$ filters. All convolutions have stride one, and all weights are randomly initialized from a Gaussian distribution with standard deviation $\sigma = 0.001$. When the training MSE of current stage stops decreasing significantly, e.g., the error decreases less than 3\% in an epoch, the network will be cascaded to a deeper network, and the training goes to the next stage. To accelerate this procedure, we insert two new layers into the network at each stage. As shown in Fig. 1, the CT-SRCNN training starts from 3 layers, and proceeds to 5 layers, 7 layers, etc.. Each new layer consists of 32 $3\times3$ filters. The small filter size of the cascaded layers is to maintain the network computation efficiency while going deeper. The new layers are inserted just before the last $5\times5$ layer. Hence, the weights of pre-existing layers could be inherited from the previous stage, and the weights of the two new layers are randomly initialized (Gaussian with $\sigma = 0.001$). We pad two zeros to each new $3\times{}3$ layer to make the size of the feature map consistent. As a result, all the stages in the cascade training have the same output size, so the training samples could be shared. 

When the network goes deeper, it is difficult for the training to converge. SRCNN \cite{dong2016image} concluded that deeper is not better as it does not show better performance with more than three layers. In VDSR \cite{kim2016accurate}, a high initial learning rate is tuned and then gradually decreased. But when using a large diverse training set (e.g., more than 30 million patches from 160,000 images), we found that a high learning rate will not work well. A potential reason is that a high  learning rate will lead to vanishing or exploding gradients. In CT-SRCNN, only a few weights are randomly initialized at each stage, so the convergence will be easy. We find that using a fixed learning rate 0.0001 without any decay for all layers  is feasible.

The trained CT-SRCNN is shown in Fig. 2, with the kernel size `9-5-3-...-3-5'. One main advantage of the deeper network is the increased receptive field which increases awareness of the context. In our implementation, the 19-layer CT-SRCNN achieves the state-of-the-art accuracy, while using much less parameters compared to other deep networks, such as VDSR \cite{kim2016accurate} or DRRN \cite{tai2017image}. We implemented the CT-SRCNN using the Caffe package. 
\begin{figure*}
\centering
\includegraphics[height=2.5cm]{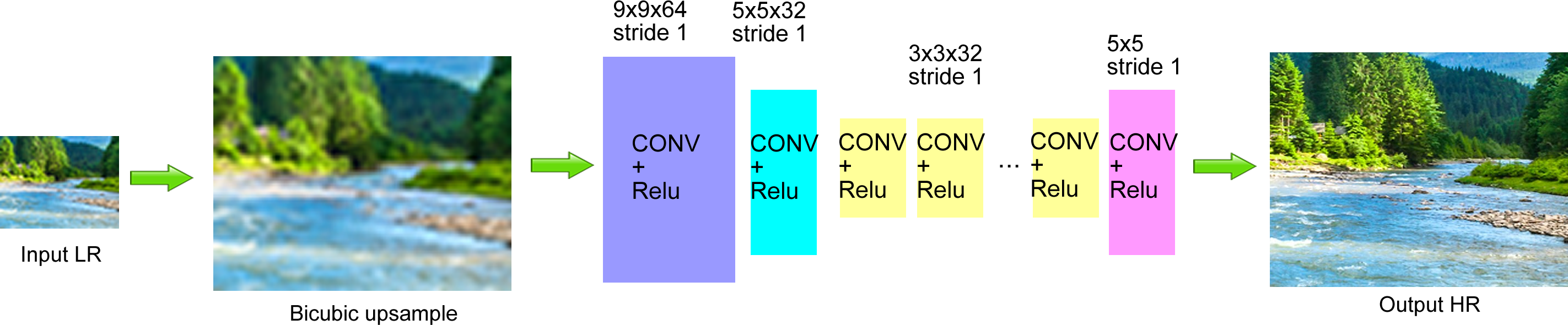}
\caption{CT-SRCNN with `9-5-3-...-3-5' network architecture.}
\label{fig:example}
\end{figure*}

\section{Improving the efficiency of the CT-SRCNN}
\subsection{One-shot trimming}
\noindent{}Most of the neural networks have redundancy. Removing such redundancy will clearly improve the efficiency. Given a $L$-layer network, assume the $i$th layer consists of $n_{i-1}$ input channels, $k_i\times{}k_i$ convolution kernel, and $n_i$ filters. The input feature map is $F_{i-1}\in{}R^{n_{i-1}\times{}h_{i-1}\times{}w_{i-1}}$, and the output feature map is $F_{i}\in{}R^{n_{i}\times{}h_{i}\times{}w_{i}}$, where $w_i$ and $h_i$ are the width and height of the feature map. Then the weights in this layer are $W_i\in{}R^{n_{i-1}\times{}k_i\times{}k_i{}\times{n_i}}$. There will be $n_{i-1}\times{}k_i\times{}k_i{}\times{n_i}\times{}w_{i}\times{}h_{i}$ multiplications when doing the convolution.

Since the major computational cost is decided by the number of the multiplications, a straightforward way to improve the efficiency is to reduce the number of the multiplications, e.g., network pruning by setting some weights in $W_i$ to 0. Although such weight pruning may reduce the compressed network size \cite{Han2016, Choi2017}, it is hard to take advantage of it in reducing computational complexity with hardware implementations by zero skipping due to the typical random pruning patterns. Another way to improve the CNN's computational efficiency is to train a slimmer network which is totally randomly initialized as in FitNets \cite{Romero2015}. But it will require the guidance from original network with heavy parameter tuning.

Instead, we propose to transform the original network to the slim network by trimming whole filters from the layers. This will reduce the network size and computational cost at the same time. Once a filter is trimmed, the adjacent layer will also be influenced, as shown in Fig. 3. If we trim a filter (red block) from the $i$th layer, $n_{i}= n_{i}-1$, some weights in the $(i+1)$th layer (red segments) will also be trimmed, resulting in kernels with reduced depth at the $(i+1)$th layer. So trimming the filter in the $i$th layer will reduce the computational cost for both the $i$th and the $i+1$th layer.
In the straightforward approach, one-shot trimming, the filters from all the layers are trimmed based on the trimming rate $\epsilon_{filters,i}, i = 1,\ldots,L$ at once. Then, the trimmed network will be fine-tuned until convergence. 

\begin{figure}
\centering
\includegraphics[height=3.6cm]{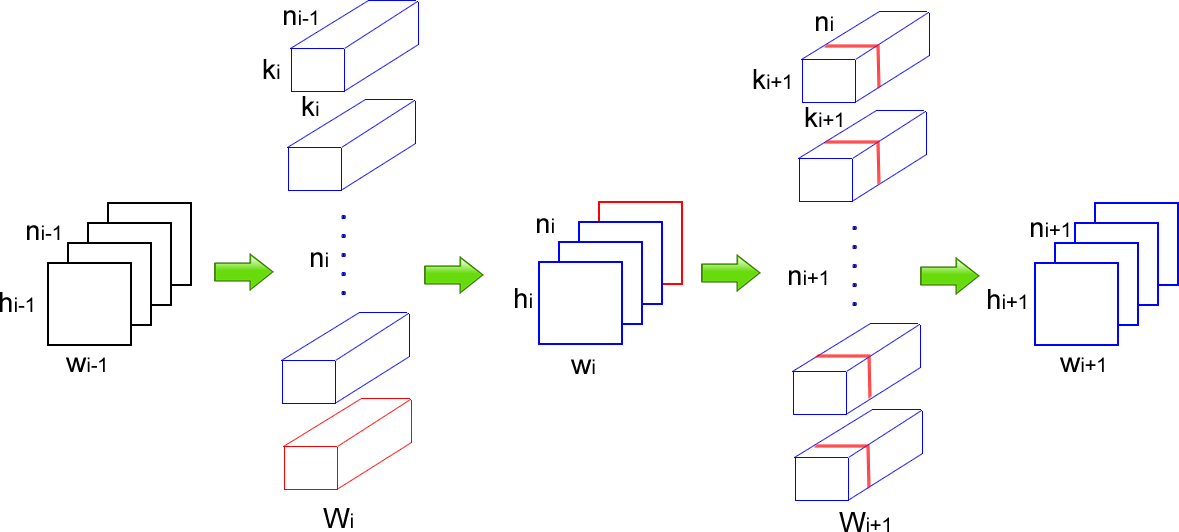}
\caption{Trimming single filter (red block) from the $i$th layer in CT-SRCNN. The $(i+1)$th layer will also be influenced.}
\label{fig:example}
\end{figure}

To decide which filter will be trimmed, we define the relative importance $R_{i,j}$ of the $j$th filter in the $i$th layer as the square sum of all the weights in this filter. As in equation (1), the $W_{i,j}$ is the weights in the $j$th filter of the $i$th layer

\begin{equation}
R_{i,j}= \sum_{w \in{}W_{i,j}}{w^2}.
\end{equation}

Then we may trim the filters with small $R_{ij}$ for each layer $i, i = 1,\ldots, L$. We know that if we trim a filter from the $i$th layer, the corresponding weights in the $(i+1)$th layer will also be trimmed, which results in $W_{i+1,j}'$. So when calculating $R_{i+1,j}$, we may either refer to the non-trimmed weights $W_{i+1,j}$ of the original network, which is called `independent trimming', or the reduced-depth kernels $W_{i+1,j}'$ which are partially trimmed as a result of trimming the previous layer, as in equation (2), which is called `greedy trimming' \cite{li2017trimming}. 

\begin{equation}
R_{i+1,j}= \sum_{w \in{}W_{i+1,j}'}{w^2}.
\end{equation}

The flowchart of applying one-shot trimming to CT-SRCNN is illustrated in Fig. 4. In the implementation, we set $\epsilon_{filters,i}=0.5, i = 1,\ldots,L$ for all $L$ layers. The output trimmed CT-SRCNN will have `32-16-16-...-16-1' filters in each layer.

\begin{figure}
\centering
\fbox{
\shortstack[l]
{Parameters \\ 
\quad $\epsilon_{filters,i}, i = 1,\ldots,L$ \quad threshold of filter trimming \\
Input: CT-SRCNN model with L layers, each layer has\\
 $n_i$ filters \\
1. Repeat for $i=1,2,\ldots,L$ \\
~~1.1 Calculate $R_{i,j},j=1,\ldots,n_i$ for all the filters in the  \\
~~ $i$th layer using (1) or (2) \\
~~1.2 Remove the $\epsilon_{filters,i}\times{}n_i$ filters from the $i$th layer\\
~~1.3 If $i<L$, remove the corresponding weights in the\\
~~$i+1$th layer \\
2. Fine-tuning and output trimmed model}} 
\caption{One-shot filter trimming of CT-SRCNN.}
\end{figure}

\subsection{Cascade trimming}
\noindent{}The network size versus the accuracy is a trade-off problem. Although the one-shot trimming may reduce the network size, the accuracy loss is also significant (0.3-0.4dB, as shown in Table 5). One reason of this loss is that for deep networks, one-shot trimming will change the network capability a lot. So it will be very difficult to retrieve the accuracy by fine-tuning the trimmed network.

Inspired by the cascade training, we proposed the cascade trimming method. Cascade trimming also consists of several stages. In each stage, we trim the filters from a few layers only. The filters in other layers are inherited from previous stage. Then the whole network is fine-tuned until convergence, without any frozen weights.  Similar to cascade training, when the MSE of current stage stops decreasing significantly, e.g., the error decreases less than 3\% in an epoch,  the trimming will move to the next stage. 

We apply the cascade trimming on the CT-SRCNNs. Two adjacent layers are trimmed in each stage for acceleration. CT-SRCNN has `64-32-32-...-32-1' filters per layer. The cascade trimming starts from the last two layers with 32 filters since there is only one filter in the final layer. Since trimming the filters in one layer will also influence the kernels at the next layer by reducing their depths, the filters are actually trimmed from 3 adjacent layers at each stage, as shown in Fig. 4. Suppose we have a $L$-layer CT-SRCNN, the first trimming stage will trim the $(L-1)$th and the $(L-2)$th layers, but also influence the last  layer. The second trimming stage will focus on the $(L-3)$th and $(L-4)$th layers, but also influence the $(L-2)$th layer, etc... The cascade trimming will terminate when all the layers are trimmed. In each trimmed layer, half of the filters will be removed. These filters are randomly selected instead of calculating small $R_{i,j}$ in the one-shot trimming. The reason is that the network capability will not be hurt much in each stage of the cascade trimming, so fine-tuning is relatively easy even with random selection. If we apply such random selection in the one-shot trimming, the fine-tuning will not converge. 

During the fine-tuning, the learning rate is fixed to 0.0001. Since the cascade trimming did not change the size of the convolution kernels and feature maps, we may use the same training samples as cascade training.

\begin{figure*}
\centering
\includegraphics[height=5cm]{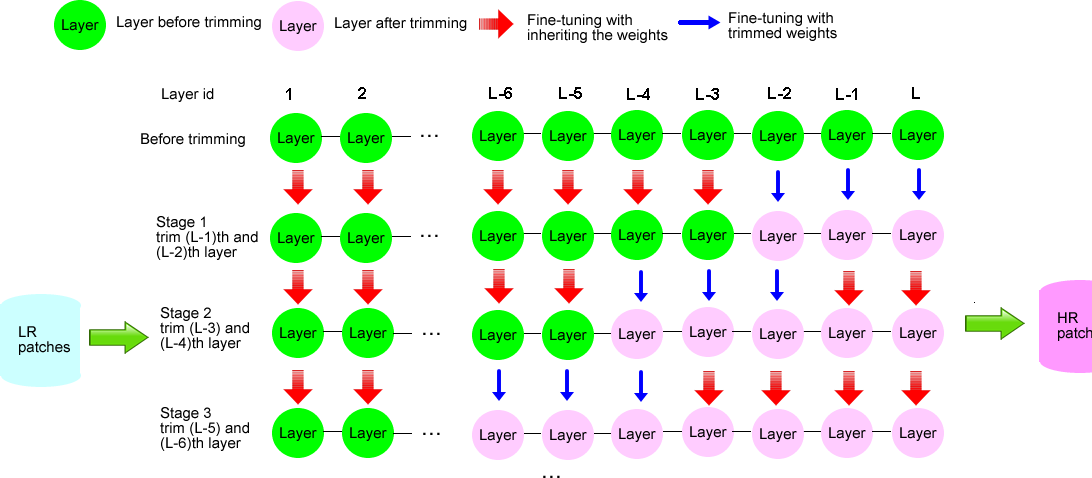}
\caption{Implementation of cascade trimming on CT-SRCNN. In each stage, the filters are trimmed from two adjacent layers, and one layer is influenced by trimming filters from its previous layer (Fig. 3). Then the whole network is fine-tuned until convergence.}
\label{fig:example}
\end{figure*}

The output trimmed CT-SRCNN has the same `32-16-16-...-16-1' architecture as the one-shot trimming. This framework is called `train-trim'. Another way to train such a network is starting by trimming and then train a deeper network. For example, trimming a 3-layer network to `32-16-1' filters, and inserting two $16\times{}3\times{}3$ filters by cascade training to make it deeper and deeper. The final network which is denoted as `trim-train' will have the same architecture as the `train-trim' network. We compare these two methodologies in section 5.3. As expected, we show that the `train-trim' performs better compared to `trim-train'. This can be attributed to the fact that in `train-trim' deeper fatter network can learn better representations due to more degrees of freedom, and then the network is made slimmer by trimming away the redundant filters.

\begin{table*}
\caption{PSNR/SSIM evaluation of CT-SRCNNs with different layers. Bold font indicates the best.}
\scriptsize
\renewcommand{\arraystretch}{1.0}
\centering
\begin{tabular}{|c|c|c|c|c|c|c|c|c|c|c|}
\hline
{} & {Scale} & {3-layer} & {5-layer} &{7-layer} &{9-layer} &{11-layer} & {13-layer} & {15-layer} & 17-layer & 19-layer \\ \hline
Para. & - & 57,184 & 75,616 & 94,048 & 112,480 & 130,912 & 149,344 & 167,776 & 186,208 & 204,640 \\ \hline
& 2 & 36.66/0.9542 & 36.88/0.9553 & 37.03/0.9567 & 37.19/0.9574 & 37.44/0.9580  & 37.61/0.9590 & 37.70/0.9599 & 37.76/0.9609 & \textbf{37.81}/\textbf{0.9618}\\
Set5 & 3 & 32.75/0.9090 & 33.13/0.9141 & 33.35/0.9169 & 33.43/0.9179 & 33.59/0.9204 & 33.76/0.9219 & 33.90/0.9230 & 33.96/0.9236 & \textbf{34.02}/\textbf{0.9244}\\ 
& 4 & 30.48/0.8628 & 30.81/0.8717 & 31.11/0.8767 & 31.22/0.8788 & 31.32/0.8825 & 31.49/0.8849 & 31.63/0.8923 & 31.74/0.8937 & \textbf{31.80}/\textbf{0.8944}\\ \hline
& 2 & 32.42/0.9063 & 32.65/0.9083 & 32.96/0.9108 & 33.12/0.9118 & 33.30/0.9128 & 33.37/0.9131 & 33.48/0.9139 & 33.57/0.9146 & \textbf{33.63}/\textbf{0.9154} \\
Set14 & 3 & 29.28/0.8209 & 29.56/0.8258 & 29.71/0.8287 & 29.75/0.8299 & 29.81/0.8307 & 29.91/0.8324 & 30.06/0.8340 & 30.14/0.8347 & \textbf{30.19}/\textbf{0.8351} \\ 
& 4 & 27.49/0.7550 & 27.77/0.7600 & 27.96/0.7642 & 28.04/0.7663 & 28.10/0.7674 & 28.20/0.7680 & 28.29/0.7695 & 28.37/0.7704 & \textbf{28.44}/\textbf{0.7717} \\ \hline
& 2 & 31.36/0.8879 & 31.47/0.8909 & 31.64/0.8934 & 31.72/0.8945 & 31.80/0.8953 & 31.87/0.8962 & 31.96/0.8970 & 32.04/0.8978 & \textbf{32.09}/\textbf{0.8982} \\
B100 & 3 & 28.41/0.7863 & 28.54/0.7912 & 28.63/0.7938 & 28.67/0.7952 & 28.69/0.7963 & 28.80/0.7980 & 28.89/0.7994 & 28.95/0.8001 & \textbf{29.00}/\textbf{0.8009} \\ 
& 4 & 26.90/0.7101 & 27.04/0.7188 & 27.14/0.7215 & 27.18/0.7229 & 27.23/0.7240 & 27.30/0.7253 & 27.41/0.7266 & 27.52/0.7276 & \textbf{27.58}/\textbf{0.7288}  \\ \hline

\end{tabular}
\end{table*}

\begin{table*}
\caption{PSNR/SSIM/time evaluation of different SR algorithms. Bold font indicates the best, blue indicates the second best. Time unit is second.}
\scriptsize
\renewcommand{\arraystretch}{1.0}
\centering
\begin{tabular}{|c|c|c|c|c|c|c|c|c|c|c|}
\hline
{Dataset} & {Scale} & bicubic & {A+ \cite{timofte2014a+}} &{SRCNN \cite{dong2016image}} & VDSR \cite{kim2016accurate} & DRCN \cite{kim2016deeply} & DRRN \cite{tai2017image} & CT-SRCNN \ \\ \hline
& 2 & 33.66/0.9299/0.00 & 36.54/0.9544/0.58  & 36.66/0.9542/\textbf{0.012} & 37.44/0.9580/0.11 & {37.59}/{0.9579}/2.05 & \textcolor[rgb]{0,0,1}{37.68}/\textcolor[rgb]{0,0,1}{0.9592}/0.78 & \textbf{37.81}/\textbf{0.9618}/\textcolor[rgb]{0,0,1}{0.029} \\
Set5 & 3 & 30.39/0.8682/0.00 & 32.58/0.9088/0.32  & 32.75/0.9090/\textbf{0.012} & 33.69/0.9215/0.11 & 33.69/0.9220/2.08 & \textcolor[rgb]{0,0,1}{33.93}/\textcolor[rgb]{0,0,1}{0.9237}/0.80 & \textbf{34.02}/\textbf{0.9244}/\textcolor[rgb]{0,0,1}{0.030} \\ 
& 4 & 28.42/0.8104/0.00 &  30.28/0.8603/0.24  & 30.48/0.8628/\textbf{0.011} & 31.37/0.8839/0.11 & {31.42}/{0.8844}/2.07 & \textcolor[rgb]{0,0,1}{31.59}/\textcolor[rgb]{0,0,1}{0.8895}/0.76 & \textbf{31.80}/\textbf{0.8944}/\textcolor[rgb]{0,0,1}{0.029} \\ \hline
& 2 & 30.24/0.8688/0.00 & 32.28/0.9056/0.86  & 32.42/0.9063/\textbf{0.021} & 32.98/{0.9118}/0.22 & {32.99}/0.9122/3.54 & \textcolor[rgb]{0,0,1}{33.29}/\textcolor[rgb]{0,0,1}{0.9139}/0.90 & \textbf{33.63}/\textbf{0.9154}/\textcolor[rgb]{0,0,1}{0.041} \\
Set14 & 3 & 27.55/0.7742/0.00 & 29.13/0.8188/0.56 & 29.28/0.8209/\textbf{0.021} & {29.82}/{0.8310}/0.23 & 29.71/0.8300/3.49 & \textcolor[rgb]{0,0,1}{29.91}/\textcolor[rgb]{0,0,1}{0.8344}/0.88 & \textbf{30.19}/\textbf{0.8351}/\textcolor[rgb]{0,0,1}{0.043} \\ 
& 4 & 26.00/0.7027/0.00 & 27.32/0.7491/0.38  & 27.49/0.7503/\textbf{0.020} & 28.07/{0.7679}/0.22 & {27.95}/0.7672/3.51 & \textcolor[rgb]{0,0,1}{28.21}/\textbf{0.7720}/0.85&  \textbf{28.44}/\textcolor[rgb]{0,0,1}{0.7717}/\textcolor[rgb]{0,0,1}{0.044} \\ \hline
& 2 & 29.56/0.8431/0.00 & 31.21/0.8863/0.59  & 31.36/0.8879/\textbf{0.012} & {31.82}/{0.8947}/0.18 & 31.80/0.8938/2.46 & \textcolor[rgb]{0,0,1}{32.01}/\textcolor[rgb]{0,0,1}{0.8969}/0.72 & \textbf{32.09}/\textbf{0.8982}/\textcolor[rgb]{0,0,1}{0.027}\\
B100 & 3 & 27.21/0.7385/0.00 & 28.29/0.7835/0.33  & 28.41/0.7863/\textbf{0.012} &{28.76}/{0.7972}/0.19 & {28.74}/0.7969/2.49 & \textcolor[rgb]{0,0,1}{28.88}/\textcolor[rgb]{0,0,1}{0.8008}/0.74 & \textbf{29.00}/\textbf{0.8009}/\textcolor[rgb]{0,0,1}{0.026} \\ 
& 4 & 25.96/0.6675/0.00 & 26.82/0.7087/0.26 & 26.90/0.7101/\textbf{0.012} & {27.35}/{0.7240}/0.18 & 27.26/0.7237/2.53 & \textcolor[rgb]{0,0,1}{27.38}/\textcolor[rgb]{0,0,1}{0.7284}/0.74 & \textbf{27.58}/\textbf{0.7288}/\textcolor[rgb]{0,0,1}{0.027} \\ \hline

\end{tabular}
\end{table*}

\section{Experiments}
\subsection{Datasets}
\noindent{}We use the Open Image Dataset \cite{googleoid} to generate the training LR and HR patches. This dataset consists of 9 million URLs to HD images that have been annotated with labels spanning over 6,000 categories. Instead of the overlarge whole dataset, we use the Thumbnail300KURL of the validation set, which consists of about 160,000 images with resolution around $640\times{}480$. These images are first bicubic downsampled to generate the LR images and then upsampled to the desire size. $33\times{}33$ LR patches and the corresponding $17\times{}17$ HR patches are cropped. With $33\times{}33$ stride, we sample more than 30 million patches for training. Other SR networks (e.g., VDSR, DRCN, DRRN) are retrained on this dataset during the evaluation.

For testing, the commonly-used benchmark image datasets, Set5 \cite{bevilacqua2012low}, Set14 \cite{zeyde2010single}, and Berkeley Segmentation Dataset test set (B100) \cite{martin2001database} are used. The Peak Signal-to-Noise-Ratio (PSNR) and Structure SImiliarity Measure (SSIM) are adopted to evaluate the performance.

\subsection{Experimental results of CT-SRCNN}
\noindent{}In Table 1, we show the PSNR/SSIM of the CT-SRCNN from 3 layers to 19 layers. It can be seen that the accuracy consistently increases along with using more layers. Although we show 19-layer CT-SRCNN at most, the accuracy could still be further improved by cascading more layers. This is consistent with `the deeper, the better'. Next, we compare CT-SRCNN-19 with other SR algorithms A+ \cite{timofte2014a+}, SRCNN \cite{dong2016image}, VDSR \cite{kim2016accurate}, DRCN \cite{kim2016deeply}, and DRRN-B1U25 \cite{tai2017image}. For fair comparison, we retrain VDSR, DRCN, and DRRN-B1U25 using Open Image Dataset. Since this dataset has more diversity compared to the 291 images, training these networks from scratch will decrease the performance around 0.4dB compared to their published results. Instead, we initalize the weights of these networks by the released pre-trained official model, and then fine-tune on Open Image Dataset. In Table 2, we observe that CT-SRCNN shows competitive accuracy. It achieves either the best or the second best in all the tests. Fig. 6 visualizes the output HR images of different SR methods. We can find that CT-SRCNN-19 is able to get sharper edge and less noise compared to VDSR, DRCN, and DRRN. The output is closer to the ground-truth.

The 19-layer CT-SRCNN only consists of 200K parameters. The network size is much smaller compared to VDSR (20 layers, 650K+ parameters), DRCN (1000K+ parameters), and DRRN (1000K+ parameters). We test all the above methods in Titan X GPU. In Table 2, it takes about 0.03ms-0.05ms per image for 19-layer CT-SRCNN, which is 4 times faster compared to VDSR, and more than 15 times faster than DRCN and DRRN. It is slower than 3-layer SRCNN because SRCNN only has 57K parameters. These results show that the cascade training is able to train deep CNNs with both high accuracy and relatively fast speed. 

\begin{figure*}
\centering
\includegraphics[height=7.5cm]{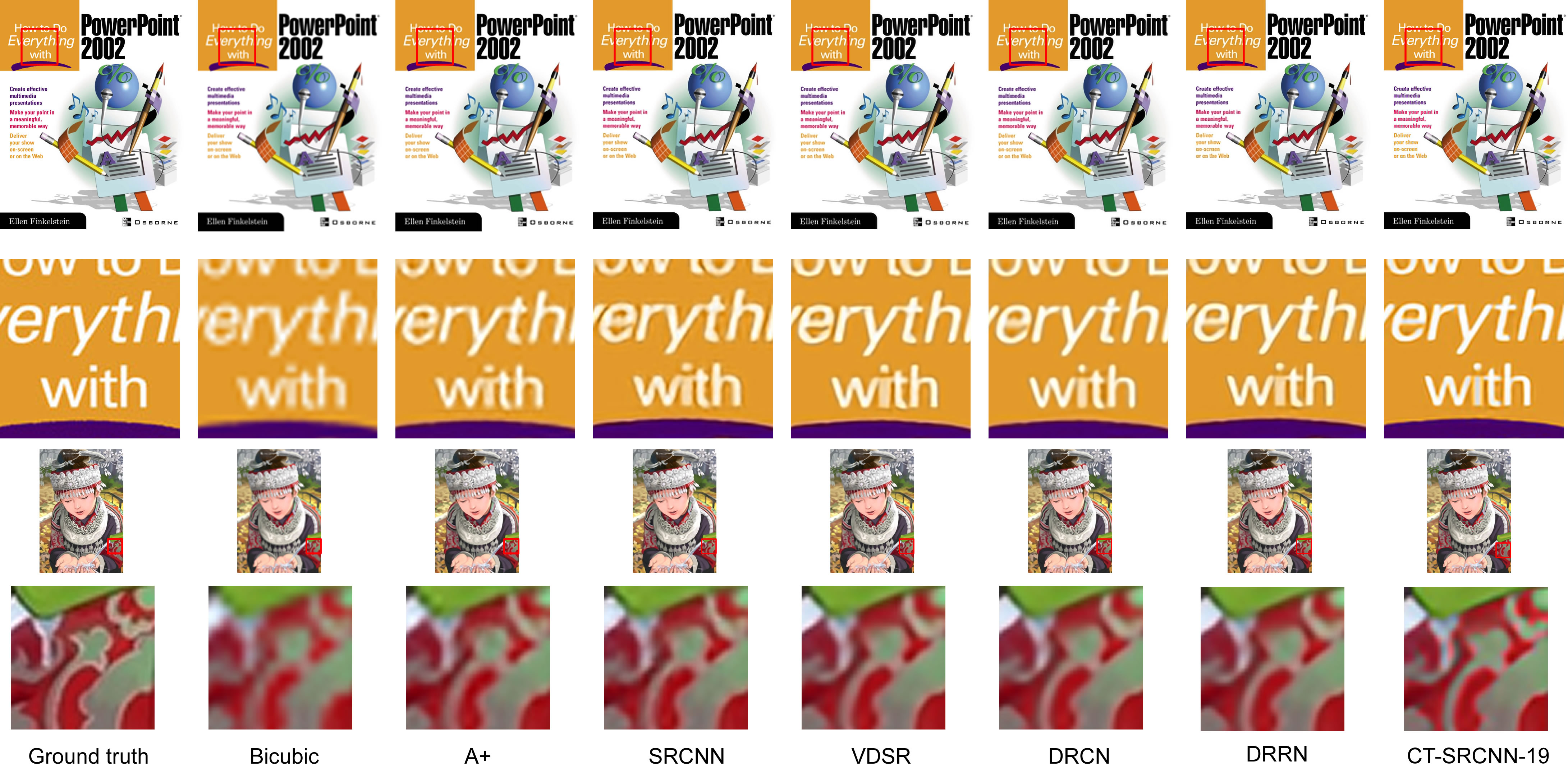}
\caption{Visualization of the outputs from different SR algorithms.}
\label{fig:example}
\end{figure*}

Moreover, we give the comparison between the CT-SRCNN and the non-cascade (one-shot) trained SRCNN with the same network architecture. For the one-shot trained SRCNN, the weight initialization and learning rate follow the method in \cite{kim2016accurate}. In Table 3, we use the 5-layer and 7-layer SRCNNs as example. It can be seen that the CT-SRCNN significantly outperforms the one-shot trained networks with the same architecture. For 7 layer, the gap is about 0.2-0.3dB. We also observe that this gap will be larger when using more layers. In Table 2, we have shown that 19-layer CT-SRCNN is clearly better than 20-layer one-shot-trained network (VDSR), while using less filters. This is due to the fact that the cascade training can be considered as a `partial-supervised initialization'. So the convergence will be easier compared to unsupervised initialization in one-shot training. Fig. 7 gives the PSNR on Set14 versus the training epochs. We find that the PSNR of CT-SRCNN increases from 3 layers to 19 layers. When the cascade training goes to the next stage, the PSNR will drop due to the random initialization of the new layers. The convergence of CT-SRCNN is better compared to one-shot trained networks.
 
 \begin{table}
\caption{Comparison between CT-SRCNN and one-shot trained SRCNNs.}
\scriptsize
\renewcommand{\arraystretch}{1.0}
\centering
\begin{tabular}{|c|c|c|c|c|c|}
\hline
{} & {Scale}  & {CT-5} &{CT-7} &{oneshot-5} &{oneshot-7}  \\ \hline
Para. & -  & 75,616 & 94,048 & 75,616 & 94,048 \\ \hline
& 2  & 36.88/0.9553 & \textbf{37.03}/\textbf{0.9567} & 36.73/0.9528 & 36.80/0.9536 \\
Set5 & 3  & 33.13/0.9141 & \textbf{33.35}/\textbf{0.9169} & 32.90/0.9112 & 32.95/0.9132\\ 
& 4  & 30.81/0.8717 & \textbf{31.11}/\textbf{0.8767} & 30.67/0.8672 & 30.75/0.8689 \\ \hline
& 2  & 32.65/0.9083 & \textbf{32.96}/\textbf{0.9108} & 32.47/0.9059 & 32.63/0.9066  \\
Set14 & 3  & 29.56/0.8258 & \textbf{29.71}/\textbf{0.8287} & 29.44/0.8232 & 29.50/0.8245 \\ 
& 4  & 27.77/0.7600 & \textbf{27.96}/\textbf{0.7642} & 27.59/0.7559 & 27.66/0.7574 \\ \hline
& 2  & 31.47/0.8909 & \textbf{31.64}/\textbf{0.8934} & 31.27/0.8880 & 31.40/0.8892 \\
B100 & 3  & 28.54/0.7912 & \textbf{28.63}/\textbf{0.7938} & 28.39/0.7872 & 28.44/0.7907 \\ 
& 4  & 27.04/0.7188 & \textbf{27.14}/\textbf{0.7215} & 26.83/0.7140 & 26.89/0.7154 \\ \hline

\end{tabular}
\end{table}

\begin{figure}
\centering
\includegraphics[height=5.5cm]{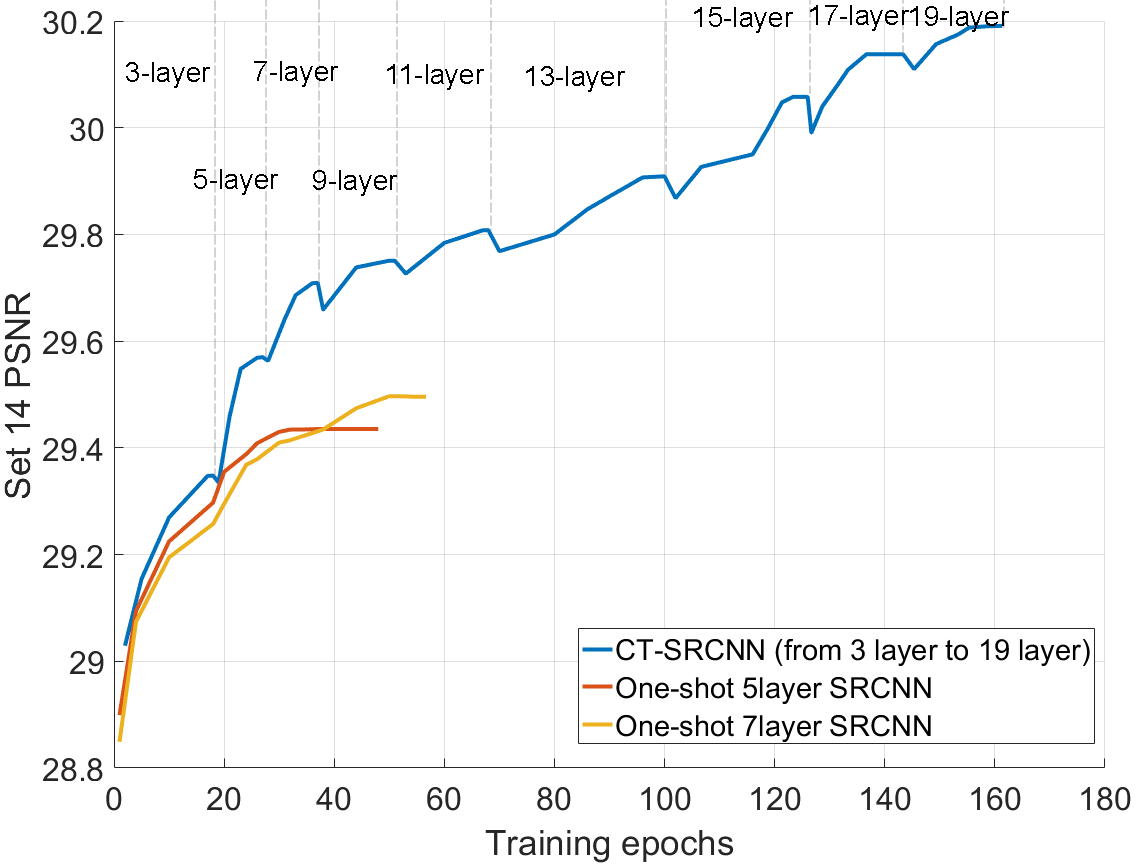}
\caption{Epochs vs. Set14 PSNR of CT-SRCNN and one-shot trained SRCNNs.}
\label{fig:example}
\end{figure}

\begin{table*}
\caption{PSNR/SSIM/time evaluation of cascade trimming on CT-SRCNN-13. Bold font indicates the best. Time unit is second.}
\scriptsize
\renewcommand{\arraystretch}{1.0}
\centering
\begin{tabular}{|c|c|c|c|c|c|c|c|c|}
\hline
 &scale& {CT-SRCNN-13} & {Trim-S1} &{Trim-S2} &{Trim-S3} &{Trim-S4} & Trim-S5 & Trim-S6 \\ \hline
Param. & - & 149,344 & 137,424 & 123,600 & 109,776 & 95,952 & 86,736 & \textbf{58,544} \\ \hline
& 2 &  \textbf{37.61}/\textbf{0.9590}/0.019 & 37.61/0.9591/0.018 & 37.60/0.9590/0.016 & 37.59/0.9589/0.015 & 37.57/0.9586/0.014 & 37.54/0.9581/0.013 & 37.50/0.9576/\textbf{0.012} \\ 
Set5& 3 & \textbf{33.76}/\textbf{0.9219}/0.019 &  33.77/0.9219/0.018 & 33.75/0.9218/0.016 & 33.74/0.9216/0.014 & 33.71/0.9213/0.014 & 33.66/0.9209/0.013 & 33.61/0.9201/\textbf{0.012} \\ 
& 4 & \textbf{31.49}/\textbf{0.8849}/0.020 & 31.49/0.8849/0.018 & 31.49/0.8848/0.017 & 31.48/0.8847/0.017 & 31.44/0.8845/0.014 & 31.40/0.8841/0.014 & 31.35/0.8836/\textbf{0.012} \\  \hline
& 2 &  \textbf{33.37}/\textbf{0.9131}/0.032 & 33.38/0.9130/0.030& 33.36/0.9130/0.028 & 33.36/0.9128/0.025 & 33.33/0.9124/0.025 & 33.29/0.9121/0.024 & 33.23/0.9115/\textbf{0.022} \\ 
Set14 & 3 & \textbf{29.91}/\textbf{0.8324}/0.034 &  29.90/0.8326/0.031 & 29.89/0.8324/0.028 & 29.88/0.8322/0.026 & 29.85/0.8318/0.025 & 29.83/0.8315/0.025 & 29.78/0.8308/\textbf{0.022} \\ 
& 4 & \textbf{28.20}/\textbf{0.7680}/0.032 & 28.20/0.7681/0.029 & 28.19/0.7679/0.027 & 28.18/0.7677/0.025 & 28.15/0.7674/0.025 & 28.13/0.7671/0.024 & 28.06/0.7664/\textbf{0.022} \\  \hline
& 2 &  \textbf{31.87}/\textbf{0.8962}/0.020 & 31.86/0.8964/0.018 & 31.86/0.8962/0.017 & 31.84/0.8960/0.015 & 31.81/0.8955/0.014 & 31.78/0.8953/0.014 & 31.76/0.8951/\textbf{0.013} \\ 
BSD & 3 & \textbf{28.80}/\textbf{0.7980}/0.021 &  28.81/0.7980/0.019 & 28.80/0.7979/0.017 & 28.78/0.7978/0.016 & 28.75/0.7974/0.014 & 28.74/0.7971/0.013 & 28.72/0.7968/\textbf{0.012} \\ 
& 4 & \textbf{27.30}/\textbf{0.7253}/0.020 & 27.30/0.7254/0.018 & 27.30/0.7251/0.016 & 27.28/0.7249/0.015 & 27.25/0.7244/0.014 & 27.23/0.7243/0.014 & 27.21/0.7238/\textbf{0.013} \\  \hline
\end{tabular}
\end{table*}

\begin{table*}
\caption{PSNR/SSIM/time evaluation of cascade trimming vs. one-shot trimming on CT-SRCNN-13. Bold font indicates the best. Time unit is second.}
\scriptsize
\renewcommand{\arraystretch}{1.0}
\centering
\begin{tabular}{|c|c|c|c|c|c|}
\hline
 &scale & {One-shot-independent} &{One-shot-greedy} &{Cascade Trimming} &{Trim-Train} \\ \hline
Param.  & - & 58,544 & 58,544 & 58,544 & 58,544 \\ \hline
& 2  & 37.20/0.9546/0.012 & 37.18/0.9549/0.012 & \textbf{37.50}/\textbf{0.9576}/0.012 &  37.33/0.9555/0.012 \\ 
Set5& 3  &  33.24/0.9161/0.012 & 33.21/0.9158/0.012& \textbf{33.61}/\textbf{0.9201}/0.012 & 33.43/0.9185/0.012\\ 
& 4  & 31.02/0.8802/0.013 & 31.05/0.8805/0.012 & \textbf{31.35}/\textbf{0.8836}/0.012 & 31.16/0.8817/0.012\\  \hline
& 2  & 32.80/0.9096/0.022& 32.76/0.9100/0.021 & \textbf{33.23}/\textbf{0.9115}/0.022 & 33.00/0.9105/0.021 \\ 
Set14 & 3  &  29.70/0.8270/0.022 & 29.68/0.8266/0.022 & \textbf{29.78}/\textbf{0.8308}/0.022 &  29.72/0.8290/0.022\\ 
& 4  & 27.91/0.7633/0.022 & 27.91/0.7639/0.022 & \textbf{28.06}/\textbf{0.7664}/0.022 & 27.97/0.7650/0.022 \\  \hline
& 2  & 31.54/0.8920/0.013 & 31.55/0.8924/0.012 & \textbf{31.76}/\textbf{0.8951}/0.013  & 31.64/0.8940/0.012 \\ 
BSD & 3  &  28.60/0.7938/0.012 & 28.61/0.7941/0.012 & \textbf{28.72}/\textbf{0.7968}/0.012 &  28.65/0.7953/0.013 \\ 
& 4  & 27.08/0.7210/0.013 & 27.09/0.7214/0.012 & \textbf{27.21}/\textbf{0.7238}/0.013 & 27.12/0.7222/0.013 \\  \hline
\end{tabular}
\end{table*}

\subsection{Experimental results of cascade trimming}
\noindent{}In Table 4, we give the experimental results of the cascade trimming. We use the 13-layer CT-SRCNN as the baseline network. Since 2 layers will be trimmed in each stage, there will be 6 trimming stages to trim the whole network. It can be seen that the first 3 trimming stages (Trim-S1 to Trim-S3) will not decrease the PSNR and SSIM very much compared to the non-trimmed model, but the network size has been reduced around 30\%. When all the 13 layers are trimmed after 6 stages (Trim-S6), the PSNR decreases only 0.15dB. At the same time, the network size is reduced to one third of the original network, and the inference time is significantly reduced.  This indicates that the cascade trimming is an effective way to deal with the efficiency-accuracy trade-off. We notice that the trimmed models consists of only 58K parameters, the accuracy is much better compared to a non-trimmed model with similar network size (3-layer SRCNN has 57K parameters). This implies that the performance of a trimmed deep network outperforms a non-trimmed network with similar size. 

In Table 5, we compare the cascade trimming with the one-shot trimming discussed in section 4.l. Due to the difference in selecting filters to be trimmed, we have two different ways, `one-shot-independent' (independent trimming) and `one-shot-greedy' (greedy trimming). It can been seen that the one-shot trimming is worse compared to cascade trimming. But it is still better than 3-layer SRCNN. So we can get the conclusion that trimming from a deeper model is always helpful, and the cascade trimming is clearly better compared to the one-shot trimming. In addition, we notice that the independent trimming achieves similar accuracy compared to the greedy trimming. We also give the comparison between the `train-trim' (Cascade Trimming) and `trim-train' (the last column), as discussed in Section 4.2. We observe that the `trim-train' is better than one-shot trimming, but worse compared to cascade trimming. This result makes sense because there will be more flexibility when learning the trimming of a cascade trained `fatter network', compared to cascade training a thinner network only. 

As summary, using the proposed cascade training and cascade trimming, we are able to train a SR network with both high accuracy and considerable efficiency.

\section{Conclusion}
\noindent{}In this paper, we presented cascade methods to train deep CNNs for super resolution achieving both high accuracy and efficiency. The cascade training ensures that the network might consistently go deeper with a relatively smaller size. The cascade trimming further reduces the network complexity, with minor loss on the discriminative power. The experimental results on benchmark image datasets show that the proposed method achieves competitive performance compared to the state-of-the-arts, but the speed is much faster. Although SRCNN has been adopted as our base SR network in this paper, we have also observed similar conclusions when applying the cascade methodologies to improve the accuracy and efficiency of other base SR networks, such as SRResNet \cite{ledig2016photo}, FSRCNN \cite{dong2016accelerating}, and ESPCN \cite{shi2016real}.
Our method could also be generalized to the CNNs for other applications, such as denoising, or image restoration.


\begin{thebibliography}{100}

\bibitem{nasrollahi2014super}
Nasrollahi, Kamal and Moeslund, Thomas.
\newblock Super-resolution: a comprehensive survey.
\newblock {\em Machine vision and applications}, 25:1423--1468, 2014.
  
\bibitem{yang2014single}
Yang, Chih-Yuan and Ma, Chao and Yang, Ming-Hsuan.
\newblock Single-image super-resolution: A benchmark.
\newblock {\em ECCV}, 372--386, 2014.

\bibitem{glasner2009super}
Glasner, Daniel and Bagon, Shai and Irani, Michal.
\newblock Image Super-resolution from a single image.
\newblock {\em CVPR}, 349--356, 2009.

\bibitem{kim2010single}
Kim, Kwang In and Kwon, Younghee.
\newblock Single-image super-resolution using sparse regression and natural image prior.
\newblock {\em TPAMI}, 32:1127--1133, 2010.

\bibitem{allebach1996edge}
Allebach, Jan and Wong, Ping Wah.
\newblock Edge-directed interpolation.
\newblock {\em ICIP}, 707--710, 1996.

\bibitem{li2001new}
Li, Xin and Orchard, Michael.
\newblock New edge-directed interpolation.
\newblock {\em TIP}, 10:1521--1527, 2001.


\bibitem{chang2004super}
Chang, Hong and Yeung, Dit-Yan and Xiong, Yimin.
\newblock Super-resolution through neighbor embedding.
\newblock {\em CVPR}, 2004.

\bibitem{bevilacqua2012low}
Bevilacqua, Marco et al.
\newblock Low-complexity single-image super-resolution based on nonnegative neighbor embedding.
\newblock {\em BMVC}, 2012.

\bibitem{timofte2013anchored}
Timofte, Radu and De Smet, Vincent and Van Gool, Luc.
\newblock Anchored neighborhood regression for fast example-based super-resolution.
\newblock {\em ICCV}, 2013.

\bibitem{timofte2014a+}
Timofte, Radu and De Smet, Vincent and Van Gool, Luc.
\newblock Anchored neighborhood regression for fast example-based super-resolution.
\newblock {\em ACCV}, 2014.

\bibitem{yang2010image}
Yang, Jianchao and Wright, John and Huang, Thomas and Ma, Yi.
\newblock Image super-resolution via sparse representation.
\newblock {\em TIP}, 19:2861-2873, 2010.

\bibitem{timofte2016seven}
Timofte, Radu and Rothe, Rasmus and Van Gool, Luc.
\newblock Seven ways to improve example-based single image super resolution.
\newblock {\em CVPR}, 1865-1873, 2016.

\bibitem{dong2016image}
Dong, Chao et al.
\newblock Image super-resolution using deep convolutional networks.
\newblock {\em TPAMI}, 38:295-307, 2016.


\bibitem{kim2016accurate}
Kim, Jiwon and Kwon Lee, Jung and Mu Lee, Kyoung.
\newblock Accurate image super-resolution using very deep convolutional networks.
\newblock {\em CVPR}, 1646-1654, 2016.

\bibitem{kim2016deeply}
Kim, Jiwon and Kwon Lee, Jung and Mu Lee, Kyoung.
\newblock Deeply-recursive convolutional network for image super-resolution.
\newblock {\em CVPR}, 1637-1645, 2016.

\bibitem{ledig2016photo}
Ledig, Christian et al.
\newblock Photo-realistic single image super-resolution using a generative adversarial network.
\newblock {\em Arxiv}, 1609.04802, 2016.

\bibitem{dahl2017pixel}
Dahl, Ryan and Norouzi, Mohammad and Shlens, Jonathon.
\newblock Pixel Recursive Super Resolution.
\newblock {\em Arxiv}, 1702.00783, 2017.

\bibitem{Romero2015}
Romero, A. and  Ballas, N. and Kahou, S. E. and Chassang, A. and Gatta, C. and Bengio, Y.
\newblock Fitnets: Hints for thin deep nets
\newblock {\em ICLR}, 2015.


\bibitem{Han2016}
Han, Song, Huizi Mao, and Dally, William.
\newblock Deep compression: Compressing deep neural networks with pruning, trained quantization and Huffman coding
\newblock {\em ICLR}, 2016.

\bibitem{Choi2017}
Choi, Yoojin and El-Khamy, Mostafa and Lee, Jungwon. 
\newblock Towards the Limit of Network Quantization
\newblock {\em ICLR}, 2017.

\bibitem{freeman2002example}
Freeman, William and Jones, Thouis and Pasztor, Egon.
\newblock Example-based super-resolution.
\newblock {\em IEEE Computer graphics and Applications}, 22:56--65, 2002.

\bibitem{huang2015single}
Huang, Jia-Bin and Singh, Abhishek and Ahuja, Narendra.
\newblock Single image super-resolution from transformed self-exemplars.
\newblock {\em CVPR}, 5197--5206, 2015.

\bibitem{tai2010super}
Tai, Yu-Wing and Liu, Shuaicheng and Brown, Michael and Lin, Stephen.
\newblock Super resolution using edge prior and single image detail synthesis.
\newblock {\em CVPR}, 2400--2407, 2010.

\bibitem{zhang2012multi}
Zhang, Kaibing and Gao, Xinbo and Tao, Dacheng and Li, Xuelong.
\newblock Multi-scale dictionary for single image super-resolution.
\newblock {\em CVPR}, 1114--1121, 2012.

\bibitem{schulter2015fast}
Schulter, Samuel and Leistner, Christian and Bischof, Horst.
\newblock Fast and accurate image upscaling with super-resolution forests.
\newblock {\em CVPR}, 3791--3799, 2015.

\bibitem{wang2015deep}
Wang, Zhaowen and Liu, Ding and Yang, Jianchao and Han, Wei and Huang, Thomas.
\newblock Deep networks for image super-resolution with sparse prior.
\newblock {\em ICCV}, 370--378, 2015.

\bibitem{Ren2017Image}
Ren, Haoyu and El-Khamy, Mostafa and Lee, Jungwon.
\newblock Image Super Resolution Based on Fusing Multiple Convolution Neural Networks.
\newblock {\em CVPRW}, 2017.

\bibitem{shi2016real}
Shi, Wenzhe et al.
\newblock Real-time single image and video super-resolution using an efficient sub-pixel convolutional neural network.
\newblock {\em CVPR}, 1874--1883, 2016.

\bibitem{dong2016accelerating}
Dong, Chao and Loy, Chen Change and Tang, Xiaoou.
\newblock Accelerating the super-resolution convolutional neural network.
\newblock {\em ECCV}, 391--407, 2016.

\bibitem{sonderby2016amortised}
S{\o}nderby, Casper et al.
\newblock Amortised MAP Inference for Image Super-resolution.
\newblock {\em ArXiv}, 1610.04490,  2016.

\bibitem{johnson2016perceptual}
Johnson, Justin and Alahi, Alexandre and Fei-Fei, Li.
\newblock Perceptual losses for real-time style transfer and super-resolution.
\newblock {\em ECCV}, 694--711, 2016.

\bibitem{li2017trimming}
Hao, Li, and Asim, Kadav, and Igor, Durdanovic, and Hanan, Samet and Hans, Graf.
\newblock Pruning filters for efficienct convnets.
\newblock {\em ICLR}, 2017.

\bibitem{zeyde2010single}
Zeyde, Roman and Elad, Michael and Protter, Matan.
\newblock On single image scale-up using sparse-representations.
\newblock {\em International conference on curves and surfaces}, 711--730, 2010.

\bibitem{martin2001database}
Martin, David et al.
\newblock A database of human segmented natural images and its application to evaluating segmentation algorithms and measuring ecological statistics.
\newblock {\em ICCV}, 416--423, 2001.

\bibitem{googleoid}
https://github.com/openimages/dataset

\bibitem{lai2017deep}
Lai, Wei-Sheng et al.
\newblock Deep Laplacian Pyramid Networks for Fast and Accurate Super-Resolution.
\newblock {\em CVPR}, 2017.

\bibitem{tai2017image}
Tai, Ying, Jian Yang, and Xiaoming Liu.
\newblock Image Super-Resolution via Deep Recursive Residual Network.
\newblock {\em CVPR}, 2017.

\bibitem{lim2017enhanced}
Lim, Bee et al.
\newblock Enhanced deep residual networks for single image super-resolution.
\newblock {\em CVPRW}, 2017.

\end{thebibliography}
\end{document}